\documentclass[11pt,a4paper]{article}
\usepackage[hyperref]{acl2020}
\usepackage{times}
\usepackage{latexsym}
\usepackage[nointegrals]{wasysym}
\usepackage{pifont}

\usepackage{microtype}

\aclfinalcopy 

\usepackage{latexsym}
\usepackage{amssymb}
\usepackage{graphicx}
\usepackage{subfig}
\usepackage{caption}
\usepackage{booktabs}
\usepackage{amsmath}
\usepackage{microtype}
\usepackage{tabularx}
\usepackage{makecell}
\usepackage{todonotes}

\newcommand{\Lagr}{\mathcal{L}}

\title{
Multidirectional Associative Optimization of \\ Function-Specific Word Representations}

\author{
Daniela Gerz$^{\spadesuit\diamondsuit}$ ~ Ivan Vuli\'c$^{\spadesuit}$ ~ Marek Rei$^\clubsuit$ ~ Roi Reichart$^\heartsuit$ ~ Anna Korhonen$^{\spadesuit}$
 \\
$^\spadesuit$Language Technology Lab, University of Cambridge\\
$^\diamondsuit$PolyAI Limited, London\\ 
$^\clubsuit$Department of Computing, Imperial College London \\
$^\heartsuit$Faculty of Industrial Engineering and Management, Technion, IIT\\
{ \small \tt \{dan,ivan\}@poly-ai.com, marek.rei@imperial.ac.uk} \\
{\small \tt roiri@ie.technion.ac.il, alk23@cam.ac.uk
}
}

\date{}

\begin{document}
\maketitle
\begin{abstract}
We present a neural framework for learning associations between interrelated groups of words such as the ones found in Subject-Verb-Object (SVO) structures. Our model induces a joint function-specific word vector space, where vectors of e.g. plausible SVO compositions lie close together. The model retains information about word group membership even in the joint space, and can thereby effectively be applied to a number of tasks reasoning over the SVO structure. We show the robustness and versatility of the proposed framework by reporting state-of-the-art results on the tasks of estimating selectional preference 
and event similarity. The results indicate that the combinations of representations learned with our task-independent model outperform task-specific architectures from prior work, while reducing the number of parameters by up to 95\%. 
\end{abstract}

\section{Introduction}
\label{sec:introduction}

Word representations are in ubiquitous usage across all areas of natural language processing (NLP) \citep{Collobert2011jmlr,Chen:2014emnlp,Melamud:2016naacl}. Standard approaches rely on the distributional hypothesis \citep{Harris1954,Schutze1993nips} and learn a \textit{single} word vector space based on word co-occurrences in large text corpora \citep{Mikolov:2013nips,Pennington:2014emnlp,Bojanowski:2017tacl}.
This purely context-based training produces general word representations that capture the broad notion of semantic relatedness and conflate a variety of possible semantic relations into a single space \citep{Hill:2015cl,Schwartz:2015conll}. However, this mono-faceted view of meaning is a well-known deficiency in NLP applications \citep{Faruqui:2016thesis,Mrksic:2017tacl} as it fails to distinguish between fine-grained word associations.

\begin{figure}[t]
    \centering
    \includegraphics[scale=0.95]{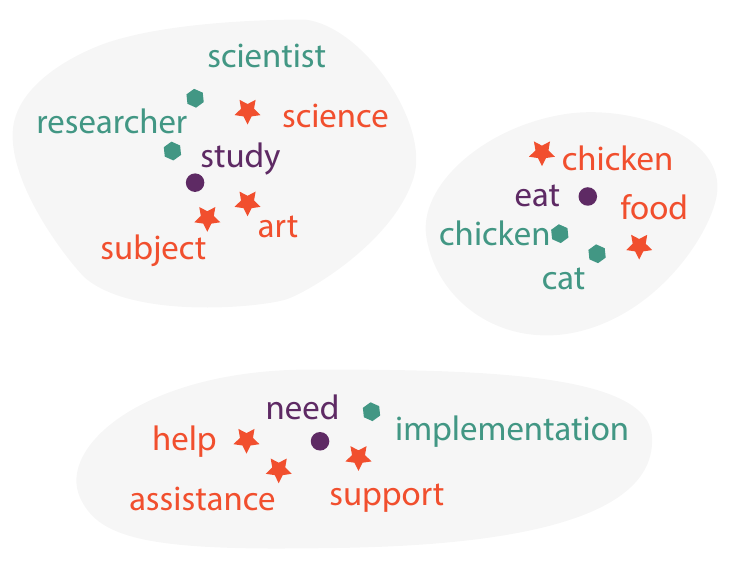}
    \caption{Illustration of three neighbourhoods in a function-specific space trained for the SVO structure (marked \Circle (S), \hexagon (V),  \ding{73}(O)). The space is optimised such that vectors for plausible SVO compositions will be close. Note that one word can have several vectors, for example \textit{chicken} can occur both as S and O. 
    }
    \label{fig:fs:motivation_nns}
\end{figure}

In this work we 
propose to learn a joint \textit{function-specific} word vector space that accounts for the different roles and functions a word can take in text. 
The space can be trained for a specific structure, such as SVO, and each word in a particular role will have a separate representation.
Vectors for plausible SVO compositions will then be optimized to lie close together, as illustrated by Figure~\ref{fig:fs:motivation_nns}.
For example, the verb vector \textit{study} will be close to plausible subject vectors \textit{researcher} or \textit{scientist} and object vectors \textit{subject} or \textit{art}. For words that can occur as either subject or object, such as \textit{chicken}, we obtain separate vectors for each role:  one for \textit{chicken} as \textit{subject} and another for \textit{chicken} as \textit{object}.
The resulting representations capture more detailed associations in addition to basic distributional similarity and can be used to construct representations for the whole SVO structure.

\begin{table}[t]
\def\arraystretch{1.1}
\setlength\tabcolsep{3pt}
\fontsize{8}{8}
\selectfont
\begin{tabularx}{1.0\linewidth}{l l}
   \cmidrule[\heavyrulewidth]{1-2}
   \textbf{Word} &  \textbf{Nearest Neighbours} \\
   \cmidrule{1-2} 
   \textbf{Subject} & {} \\
   {memory} &  {dream, feeling, shadow, sense, moment, consciousness}  \\
   {country} &  {state, nation, britain, china, uk, europe, government}  \\
   {student} &  {pupil, participant, learner, candidate, trainee, child}  \\
   \cmidrule{1-2}
   \textbf{Verb} & {} \\
   {see} &  {saw, view, expect, watch, notice, witness}  \\
   {eat} &  {drink, consume, smoke, lick, swallow, cook, ingest}  \\
   {avoid} &  {eliminate, minimise, anticipate, overcome, escape}  \\
   \cmidrule{1-2}
   \textbf{Object} & {} \\
   {virus} &  {bacteria, infection, disease, worm, mutation, antibody}  \\
   {beer} &  {ale, drink, pint, coffee, tea, wine, soup, champagne}  \\
   \cmidrule{1-2}
   \multicolumn{2}{l}{\textbf{Joint SVO}} \\
   {study (V)} & {researcher (S), scientist (S), subject (O), art (O)} \\
   {eat (V)} & {food (O), cat (S), dog (S)} \\
   {need (V)} & {help (O), implementation (S), support (O)} \\
   \cmidrule[\heavyrulewidth]{1-2}
\end{tabularx}
\label{tab:nns}
\caption{Nearest neighbours in a function-specific space trained for the SVO structure. In the \textit{Joint SVO} space (bottom) we show nearest neighbors for verbs (V) from the two other subspaces (O and S).}
\end{table}

To validate the effectiveness of our representation framework in language applications, we focus on modeling a prominent linguistic phenomenon: a general model of \textit{who does what to whom} \citep{Gell2011origin}. In language, this event understanding information is typically captured by the SVO structures and, according to the cognitive science literature, is well aligned with how humans process sentences \citep{McRae1997,Mcrae1998,Grefenstette2011a,Kartsaklis2014}; it reflects the likely distinct storage and processing of objects (typically nouns) and actions (typically verbs) in the brain \citep{Caramazza1991,Damasio1993}.

The quantitative results are reported on two established test sets for compositional event similarity \citep{Grefenstette2011a,Kartsaklis2014}. This task requires reasoning over SVO structures and quantifies the plausibility of the SVO combinations by scoring them against human judgments. We report consistent gains over established word representation methods, as well as over two recent tensor-based architectures \citep{Tilk2016,Weber2018} which are designed specifically for solving the event similarity task.

Furthermore, we investigate the generality of our approach by also applying it to other types of structures.
We conduct additional experiments in a 4-role setting, where indirect objects are also modeled, along with a \textit{selectional preference} evaluation of 2-role SV and VO relationships \citep{Chambers2010,VanDeCruys2014emnlp}, yielding the highest scores on several established benchmarks.

\section{Background and Motivation}
\label{sec:background}

\noindent \textbf{Representation Learning.}
Standard word representation models such as skip-gram negative sampling (SGNS) \citep{Mikolov:2013nips,Mikolov:2013iclr}, Glove \citep{Pennington:2014emnlp}, or FastText \citep{Bojanowski:2017tacl} induce a single word embedding space capturing broad semantic relatedness \citep{Hill:2015cl}. For instance, SGNS makes use of two vector spaces for this purpose, which are referred to as $A_{w}$ and  $A_{c}$.
SGNS has been shown to approximately correspond to factorising a matrix $M = A_{w} A_{c}^T$, where elements in $M$ represent the co-occurrence strengths between \textit{words} and their \textit{context} words \citep{Levy2014}. Both matrices represent the same vocabulary: therefore, only one of them is needed in practice to represent each word. Typically only $A_w$ is used while $A_c$ is discarded, or the two vector spaces are averaged to produce the final space. 

\citet{levy2014dependency} used dependency-based contexts, resulting in two separate vector spaces; however, the relation types were embedded into the vocabulary and the model was trained only in one direction. \citet{Camacho:2019acl} proposed to learn separate sets of relation vectors in addition to standard word vectors and showed that such relation vectors encode knowledge that is often complementary to what is coded in word vectors.
\citet{rei2018scoring} and \citet{Vulic2018} described related task-dependent neural nets for mapping word embeddings into relation-specific spaces for scoring lexical entailment. In this work, we propose a \textit{task-independent} approach and extend it to work with a variable number of relations.

\bigskip
\noindent \textbf{Neuroscience.} Theories from cognitive linguistics and neuroscience reveal that single-space representation models fail to adequately reflect the organisation of semantic concepts in the human brain (i.e., \textit{semantic memory}): there seems to be no single semantic system indifferent to modalities or categories in the brain \citep{Riddoch1988}. Recent fMRI studies strongly support this proposition and suggest that semantic memory is in fact a widely distributed neural network \citep{Davies2009,Huth2012,Pascual2015,Rice2015,deHeer2017}, where sub-networks might activate selectively or more strongly for a particular function such as modality-specific or category-specific semantics (such as objects/actions, abstract/concrete, animate/inanimate, animals, fruits/vegetables, colours, body parts, countries, flowers, etc.) \citep{Warrington1975,Warrington1987,McCarthy1988}. This indicates a \textit{function-specific} division of lower-level semantic processing. Single-space distributional word models have been found to partially correlate to these distributed brain activity patterns \citep{Mitchell2008,Huth2012,Huth2016,Anderson:2017tacl}, but fail to explain the full spectrum of fine-grained word associations humans are able to make. Our work has been partly inspired by this literature.

\bigskip
\noindent \textbf{Compositional Distributional Semantics.} Partially motivated by similar observations, prior work frequently employs tensor-based methods for composing separate tensor spaces \citep{Coecke2010}: there, syntactic categories are often represented by tensors of different orders based on assumptions on their relations. One fundamental difference is made between atomic types (e.g., nouns) versus compositional types (e.g., verbs). Atomic types are seen as standalone: their meaning is independent from other types. On the other hand, verbs are compositional as they rely on their subjects and objects for their exact meaning. Due to this added complexity, the compositional types are often represented with more parameters than the atomic types, e.g., with a matrix instead of a vector. The goal is then to compose constituents into a semantic representation which is independent of the underlying grammatical structure. Therefore, a large body of prior work is concerned with finding appropriate composition functions \citep{Grefenstette2011a,Grefenstette2011b,KartsaklisSP12,Milajevs2014} to be applied on top of word representations. Since this approach represents different syntactic structures with tensors of varying dimensions, comparing syntactic constructs is not straightforward. This compositional approach thus struggles with transferring the learned knowledge to downstream tasks. 

State-of-the-art compositional models \citep{Tilk2016,Weber2018} combine similar tensor-based approaches with neural training, leading to task-specific compositional solutions. While effective for a task at hand, the resulting models rely on a large number of parameters and are not robust: we observe deteriorated performance on other related compositional tasks, as shown in Section \ref{sec:results}. 


\bigskip
\noindent \textbf{Multivariable (SVO) Structures in NLP.} Modeling SVO-s is important for tasks such as compositional \textit{event similarity} using all three variables, and \textit{thematic fit} modeling based on \textit{SV} and \textit{VO} associations separately. Traditional solutions are typically based on clustering of word co-occurrence counts from a large corpus \citep{Baroni:2010,Greenberg2015a,Greenberg2015b,Sayeed2016,Emerson:2016repl}. More recent solutions combine neural networks with tensor-based methods. \citet{VanDeCruys2014emnlp} present a feedforward neural net trained to score compositions of both two and three groups with a max-margin loss. \citet{Grefenstette2011a,Grefenstette2011b,Kartsaklis2014,Milajevs2014,Edelstein2016} employ tensor compositions on standard single-space word vectors. \citet{Hashimoto2016} discern compositional and non-compositional phrase embeddings starting from HPSG-parsed data.  



\bigskip
\noindent \textbf{Objectives.} 
We propose to induce function-specific vector spaces which enable a better model of associations between concepts and consequently improved event representations by encoding the relevant information directly into the parameters for each word during training. Word vectors offer several advantages over tensors: a large reduction in parameters and fixed dimensionality across concepts. This facilitates their reuse and transfer across different tasks. For this reason, we find our multidirectional training to deliver good performance: the same function-specific vector space achieves state-of-the-art scores across multiple related tasks, previously held by task-specific models. 


\section{Function-specific Representation Space}
\label{sec:model}


\begin{figure*}[t]
\centering
\subfloat[Predicting $n\rightarrow 1$]{%
  \includegraphics[width=5.45cm]{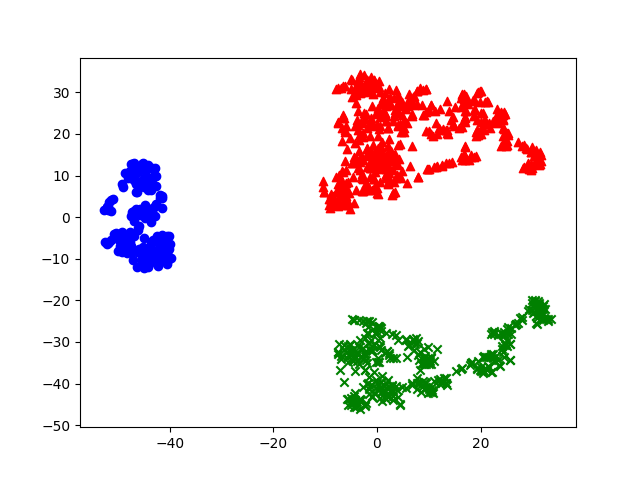}%
  \label{fig:directionality:async-input}%
}
\subfloat[Predicting $1\rightarrow n$]{%
  \includegraphics[width=5.45cm]{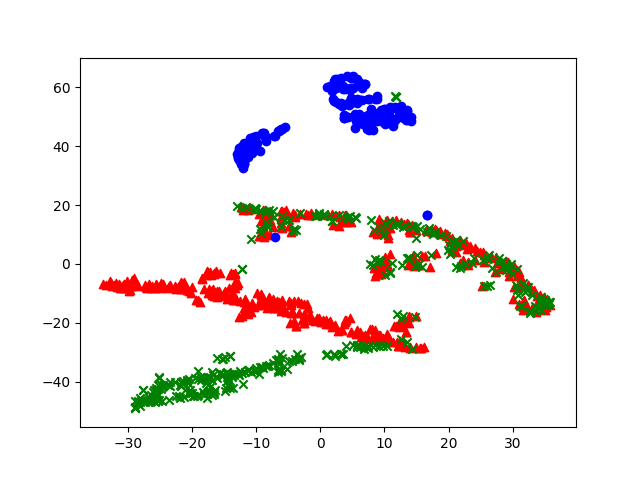}%
  \label{fig:directionality:async-output}%
}
\subfloat[Our multidirectional approach]{%
  \includegraphics[width=5.45cm]{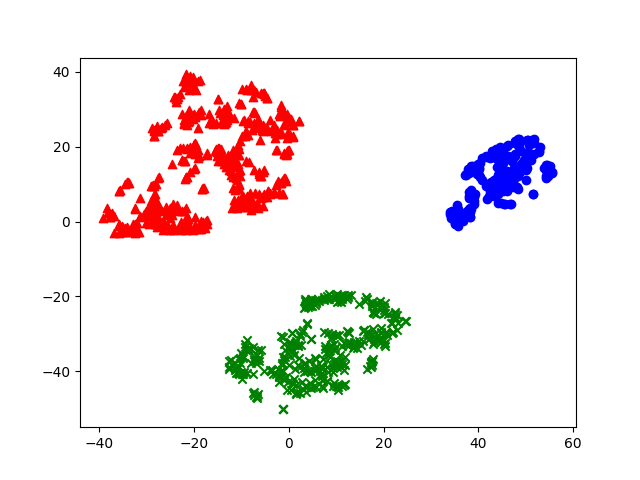}%
  \label{fig:directionality:sync}%
}
\caption{The directionality of prediction in neural models is important. Representations can be of varying quality depending on whether they are induced at the input or output side of the model. Our multidirectional approach resolves this problem by training on shared representations in all directions.}
\label{fig:directionality:visual}
\end{figure*}

Our goal is to model the mutual associations (co-occurrences) between $N$ groups of words, where each group represents a particular role, such as \textit{subject} or \textit{object} in an SVO structure.
We induce an embedding matrix $ \mathbb{R} ^ {|V_i| \times d}$ for every group $i=1,\ldots,N$, where $|V_i|$ corresponds to the vocabulary size of the $i$-th group and the group vocabularies can partially overlap. For consistency, the vector dimensionality $d$ is kept equal across all variables.

\bigskip
\noindent \textbf{Multiple Groups.} 
Without loss of generality we present a model which creates a function-specific vector space for $N=3$ groups, referring to those groups as $A$, $B$, and $C$. Note that the model is not limited to this setup, as we show later in Section \ref{sec:results}. $A$, $B$ and $C$ might be interrelated phenomena, and we aim for a model which can reliably score the plausibility of combining three vectors ($\vec{A}$,$\vec{B}$,$\vec{C}$) taken from this space.
In addition to the full joint prediction, we aim for any two vector combinations ($\vec{A}\vec{B}$, $\vec{B}\vec{C}$, $\vec{C}\vec{A}$) to have plausible scores of their own. Observing relations between words inside single-group subspaces ($A$, $B$, or $C$) is another desirable feature.

\bigskip
\noindent \textbf{Directionality.}
To design a solution with the necessary properties, we first need to consider the influence of \textit{prediction directionality} in representation learning. A representation model such as SGNS \citep{Mikolov:2013iclr,Mikolov:2013nips} learns two vectors for each word in one large vocabulary: one vector on the input side (word vector), another on the output side (context vector), with only the input word vectors being commonly used \cite{Levy2014}. Here, we require several distinct vocabularies (i.e., three, one each for group $A$, $B$, and $C$). Instead of context vectors, we train the model to predict words from another group, hence directionality is an important consideration. 

We find that prediction directionality has a strong impact on the quality of the induced representations, and illustrate this effect on an example that is skewed extremely to one side: an \textit{n:1} assignment case. Let us assume data of two groups, where each word of group $A_{1}$ is assigned to exactly one of three clusters in group $B_{3}$. We expect a function-specific word vector space customised for this purpose to show three clearly separated clusters. Figure~\ref{fig:directionality:visual} visualises obtained representations.\footnote{We train on 10K randomly selected German nouns ($A_{1}$) and their corresponding noun gender ($B_{3}$) from a German-English dictionary obtained from \texttt{dict.cc}, and train a 25-dim model for 24 epochs. Points in the figures show 1K words which were randomly selected from the 10K training vocabulary. The embedding spaces have been mapped to 2D with tSNE \citep{tsne:2012}.} Figure \ref{fig:directionality:async-input} plots the vector spaces when we use words on the input side of the model and predict the cluster: $A_{1} \rightarrow B_{3}$; this can be seen as \textit{n:1} assignment. In the opposite direction ($B_{3} \rightarrow A_{1}$, \textit{1:n} assignment) we do not observe the same trends (Figure \ref{fig:directionality:async-output}).

Representations for other and more complex phenomena suffer from the same issue. For example, the verb \textit{eat} can take many arguments corresponding to various food items such as \textit{pizza}, \textit{beans}, or \textit{kimchi}. A more specific verb such as \textit{embark} might take only a few arguments such as \textit{journey}, whereas \textit{journey} might be fairly general and can co-occur with many other verbs themselves. We thus effectively deal with an \textit{n:m} assignment case, which might be inclined towards \textit{1:n} or \textit{n:1} entirely depending on the words in question. Therefore, it is unclear whether one should rather construct a model predicting \textit{verb} $\rightarrow$ \textit{object} or \textit{object} $\rightarrow$ \textit{verb}.
We resolve this fundamental design question by training representations in a \textit{multidirectional} way with a \textit{joint loss} function.
Figure~\ref{fig:directionality:sync} shows how this method learns accurately clustered representations without having to make directionality assumptions.



\section{Multidirectional Synchronous Representation Learning}
\label{ss:sync}

The multidirectional neural representation learning model takes a list of $N$ groups of words $(G_1,G_2,\ldots,G_N)$, factorises it into all possible ``group-to-group'' sub-models, and trains them jointly by combining objectives based on skip-gram negative sampling \cite{Mikolov:2013iclr,Mikolov:2013nips}. We learn a joint function-specific word vector space by using sub-networks that each consume one group $G_i$ on the input side and predict words from a second group $G_j$ on the output side, $i,j=1,2\ldots,N; i \neq j$. All sub-network losses are tied into a single joint loss and all groups $G_1,\ldots,G_n$ are shared between the sub-networks. 

\bigskip
\noindent \textbf{Sub-Network Architecture.} 
We first factorise groups into sub-networks, representing all possible directions of prediction. Two groups would lead to two sub-networks ${A}\rightarrow{B}$ and ${B}\rightarrow{A}$; three groups lead to six sub-networks.

Similar to \cite{Mikolov:2013iclr,Mikolov:2013nips}, we calculate the dot-product between two word vectors to quantify their association. For instance, the sub-network ${A}\rightarrow{B}$ computes its prediction:
\begin{equation}
P_{A\rightarrow B} = \sigma(\vec{a} \cdot B^T_e + \vec{b}_{ab})
\end{equation}

\noindent where $\vec{a}$ is a word vector from the input group $A$, $B_e$ is the word embedding matrix for the target group $B$, $\vec{b}_{ab}$ is a bias vector, and $\sigma$ is the sigmoid function. The loss of each sub-network is computed using cross-entropy between this prediction and the correct labels:
\begin{equation}
\Lagr_{A \rightarrow B} = cross\_entropy(P_{A\rightarrow B}, L_{A\rightarrow B}).
\end{equation}

\noindent $L_{A\rightarrow B}$ are one-hot vectors corresponding to the correct predictions. We leave experiments with more sophisticated sub-networks for future work.


\bigskip
\noindent \textbf{Synchronous Joint Training.} We integrate all sub-networks into one joint model via two following mechanisms: 

\textbf{(1) Shared Parameters.} The three embedding matrices referring to groups $A$, $B$ and $C$ are shared across all sub-networks. That is, we train one matrix per group, regardless of whether it is being employed at the input or the output side of any sub-network. This leads to a substantial reduction in the model size. For example, with a vocabulary of $50,000$ words and $25$-dimensional vectors 
we work only with 1.35M parameters. Comparable models for the same tasks are trained with much larger sets of parameters: 26M or even up to 179M when not factorised \citep{Tilk2016}. Our modeling approach thus can achieve more that 95\% reduction in the number of parameters. 

\textbf{(2) Joint Loss.} We also train all sub-networks with a single joint loss and a single backward pass. We refer to this manner of joining the losses as \textit{synchronous}: it synchronises the backward pass of all sub-networks. This could also be seen as a form of multi-task learning, where each sub-network optimises the shared parameters for a different task \citep{Ruder:2017mtask}. In practice, we perform a forward pass in each direction separately, then join all sub-network cross-entropy losses and backpropagate this joint loss through all sub-networks in order to update the parameters. The different losses are combined using addition: 
\begin{equation}
\Lagr = \sum_\mu \Lagr_\mu
\end{equation}

\noindent where $\mu$ iterates over all the possible sub-networks, $\Lagr_\mu$ is the corresponding loss from one network, and $\Lagr$ the overall joint loss. 

When focusing on the SVO structures, the model will learn one joint space for the three groups of embeddings (one for $S$, $V$ and $O$).
The 6 sub-networks all share parameters and optimization is performed using the joint loss:

\begin{equation}
\begin{split}
    \Lagr = &\Lagr_{S \rightarrow V} + \Lagr_{V \rightarrow S} + \Lagr_{V \rightarrow O} \\&+ \Lagr_{O \rightarrow V} + \Lagr_{S \rightarrow O} + \Lagr_{O \rightarrow S}
\end{split}
\end{equation}

\noindent The vectors from the induced function-specific space can then be composed by standard composition functions \citep{Milajevs2014} to yield \textit{event representations} \citep{Weber2018}, that is, representations for the full SVO structure.

\section{Evaluation}
\label{sec:evaluation}



\textbf{Preliminary Task: Pseudo-Disambiguation.}
\label{subsec:binary_classification}
In the first evaluation, we adopt a standard \textit{pseudo-disambiguation} task from the selectional preference literature \citep{Rooth:1999acl,Bergsma:2008emnlp,Erk2010,Chambers2010,VanDeCruys2014emnlp}. For the three-group (S-V-O) case, the task is to score a \textit{true} triplet (i.e., the (S-V-O) structure attested in the corpus) above all \textit{corrupted} triplets (S-V'-O), (S'-V-O), (S-V-O'), where S', V' and O' denote subjects and objects randomly drawn from their respective vocabularies. Similarly, for the two-group setting, the task is to express a higher preference towards the attested pairs (V-O) or (S-V) over corrupted pairs (V-O') or (S'-V). We report accuracy scores, i.e., we count all items where \textit{score(true)} $>$ \textit{score(corrupted)}. 

This simple pseudo-disambiguation task serves as a preliminary sanity check: it can be easily applied to a variety of training conditions with different variables. However, as pointed out by \citet{Chambers2010}, the performance on this task is strongly influenced by a number of factors such as vocabulary size and the procedure for constructing corrupted examples. Therefore, we additionally evaluate our models on a number of other established datasets \citep{Sayeed2016}.


\bigskip
\noindent \textbf{Event Similarity (3 Variables: SVO).}
A standard task to measure the plausibility of SVO structures (i.e., \textit{events}) is \textit{event similarity} \citep{Grefenstette2011a,Weber2018}: the goal is to score similarity between SVO triplet pairs and correlate the similarity scores to human-elicited similarity judgements. Robust and  flexible  event  representations  are  important  to many  core  areas  in  language  understanding such as script learning, narrative generation, and discourse understanding \citep{Chambers:2009acl,Pichotta:2016aaai,Modi:2016conll,Weber2018}. 
We evaluate event similarity on two benchmarking data sets: \textbf{GS199} \citep{Grefenstette2011a} and \textbf{KS108} \citep{Kartsaklis2014}. GS199 contains 199 pairs of $SVO$ triplets/events. In the GS199 data set only the $V$ is varied, while $S$ and $O$ are fixed in the pair: this evaluation prevents the model from relying only on simple lexical overlap for similarity computation.\footnote{For instance, the phrases \textit{'people run company'} and  \textit{'people operate company'} have a high similarity score of $6.53$, whereas \textit{'river meet sea'} and \textit{'river satisfy sea'} have been given a low score of $1.84$.} KS108 contains 108 event pairs for the same task, but is specifically constructed without any lexical overlap between the events in each pair. 

For this task function-specific representations are composed into a single \textit{event representation/vector}. Following prior work, we compare cosine similarity of event vectors to averaged human scores and report Spearman's $\rho$ correlation with human scores. We compose the function-specific word vectors into event vectors using simple addition and multiplication, as well as more sophisticated compositions from prior work \cite[\textit{inter alia}]{Milajevs2014}. The summary is provided in Table~\ref{tab:compose}.

\begin{table}[t]
\def\arraystretch{1.0}
\centering
\small
\begin{tabularx}{1.0\linewidth}{ll X}
   \textbf{Data set}  & \textbf{Train}  & \textbf{Test}\\
   \midrule
   \textbf{SVO+iO} & {187K} & {15K}  \\
   \midrule
   \textbf{SVO} & {22M} & {214K}  \\
   \cmidrule{2-3}
   \textbf{} & \textbf{Vocab size} & \textbf{Freq.} \\
   {S} & 22K & {people,one,company,student} \\
   {V} & 5K & {have,take,include,provide} \\
   {O} & 15K & {place,information,way,number} \\
   \midrule
   \textbf{SV} & {69M} & {232K} \\
   \cmidrule{2-3}
   \textbf{} & \textbf{Vocab size} & \textbf{Freq.} \\
   {S} & {45K} & {people,what,one,these} \\ 
   {V} & {19K} & {be,have,say,take,go} \\
   \midrule
   \textbf{VO} & {84M} & {240K} \\
   \cmidrule{2-3}
   \textbf{} & \textbf{Vocab size} & \textbf{Freq.} \\
   {V} & {9K} & {have,take,use,make,provide} \\
   {O} & {32K} & {information,time,service} \\
\bottomrule
\end{tabularx}
\caption{Training data statistics.}
\label{tab:data_stats}
\end{table}

\begin{table}[t]
\def\arraystretch{1.0}
\footnotesize
\begin{tabularx}{1.0\linewidth}{Xc}
   \textbf{Model} &  \textbf{Accuracy} \\
   \midrule
   \textbf{4 Variables} &  \textbf{}  \\
   {SVO+iO} & {0.950} \\
   \midrule
   \textbf{3 Variables: SVO} &  \textbf{}  \\
   {\citet{VanDeCruys2009}} &  {0.874}  \\
   {\citet{VanDeCruys2014emnlp} } &  {0.889}  \\
   {\citet{Tilk2016} $\Diamond$ } &  {0.937}  \\
   {Ours } &  {0.943}  \\
   \midrule
   \textbf{2 Variables} &  \textbf{}  \\
   {\citet{Rooth:1999acl}} & {0.720} \\
   {\citet{Erk2010}} & {0.887} \\
   {\citet{VanDeCruys2014emnlp}} & {0.880} \\
   {Ours: SV} & {0.960} \\
   {Ours: VO} & {0.972} \\
\bottomrule
\end{tabularx}
\caption{Accuracy scores on the pseudo disambiguation task. $\Diamond$ indicates our reimplementation.}
\label{tab:results_true_vs_corrupted}
\end{table}

\bigskip
\noindent \textbf{Thematic-Fit Evaluation (2 Variables: SV and VO).}
Similarly to the 3-group setup, we also evaluate the plausibility of $SV$ and $VO$ pairs separately in the 2-group setup. The selectional preference evaluation \citep{Sayeed2016}, also referred to as \textit{thematic-fit}, quantifies the extent to which a noun fulfils the selectional preference of a verb
given a role (i.e., agent:S, or patient:O) \citep{McRae1997}. We evaluate our 2-group function-specific spaces on two standard benchmarks: \textbf{1)} \textbf{MST1444} \citep{Mcrae1998} contains 1,444 word pairs where humans provided thematic fit ratings on a scale from 1 to 7 for each noun to score the plausibility of the noun taking the agent role, and also taking the patient role.\footnote{Using an example from \citet{Sayeed2016}, the human participants were asked ``how common is it for a \{snake, monster, baby, cat\} to frighten someone/something'' (agent role) as opposed to ``how common is it for a \{snake, monster, baby, cat\} to be frightened by someone/something'' (patient role).} \textbf{2) \textbf{PADO414}} \citep{Pado2007} is similar to MST1444, containing 414 pairs with human thematic fit ratings,  where role-filling nouns were selected to reflect a wide distribution of scores for each verb. 
We compute plausibility by simply taking the cosine similarity between the verb vector (from the $V$ space) and the noun vector from the appropriate function-specific space ($S$ space for agents; $O$ space for patients). We again report Spearman's $\rho$ correlation scores.




\label{sec:experiments}


\bigskip
\noindent \textbf{Training Data.} We parse the ukWaC corpus \citep{Baroni2009wacky} and the British National Corpus (BNC) \citep{Leech1992bnc} using the Stanford Parser with Universal Dependencies v1.4 \citep{Chen:2014emnlp,Nivre2016universal} and extract co-occurring subjects, verbs and objects. All words are lowercased and lemmatised, and tuples containing non-alphanumeric characters are excluded. We also remove tuples with (highly frequent) pronouns as subjects, and filter out training examples containing words with frequency lower than 50. After preprocessing, the final training corpus comprises 22M SVO triplets in total. Table \ref{tab:data_stats} additionally shows training data statistics when training in the 2-group setup (SV and VO) and in the 4-group setup (when adding indirect objects: SVO+iO). We report the number of examples in training and test sets, as well as vocabulary sizes and most frequent words across different categories.

\bigskip
\noindent \textbf{Hyperparameters.} We train with batch size 128, and use Adam for optimisation \citep{Kingma2014adam} with a learning rate 0.001. All gradients are clipped to a maximum norm of 5.0. All models were trained with the same fixed random seed. We train $25$-dimensional vectors for all setups (2/3/4 groups), and we additionally train $100$-dimensional vectors for the 3-group (SVO) setup.





\section{Results and Analysis}
\label{sec:results}

\textbf{Pseudo-Disambiguation.}
\label{subsec:truevscorr}
Accuracy scores on the pseudo-disambiguation task in the 2/3/4-group setups are summarised in Table~\ref{tab:results_true_vs_corrupted}.\footnote{We also provide baseline scores taken from prior work, but the reader should be aware that the scores may not be directly comparable due to the dependence of this evaluation on factors such as vocabulary size and sampling of corrupted examples \citep{Chambers2010,Sayeed2016}.} We find consistently high pseudo-disambiguation scores ($>$0.94) across all setups. In a more detailed analysis, we find especially the prediction accuracy of verbs to be high: we report accuracy of $96.9\%$ for the 3-group SVO model. The vocabulary size for verbs is typically lowest (see Table~\ref{tab:data_stats}), which presumably makes predictions into this direction easier. In summary, as mentioned in Section \ref{sec:evaluation}, this initial evaluation already suggests that our model is able to capture associations between interrelated groups which are instrumental to modeling SVO structures and composing event representations.



\begin{table}[t]
\scriptsize
\def\arraystretch{1.0}
\begin{tabularx}{\linewidth}{lll}
  \textbf{Composition} &  \textbf{Reference} & \textbf{Formula} \\
  \cmidrule(lr){1-3}
  {Verb only} & {\newcite{Milajevs2014}}   & $\vec{V}$ \\
  {Addition} & \newcite{MitchellLapata2008}  & $\vec{S} + \vec{V} + \vec{O}$  \\
  {Copy Object} & \newcite{KartsaklisSP12}  & $\vec{S} \odot (\vec{V}\times\vec{O})$  \\
  {Concat} & {\newcite{Edelstein2016}} & [$\vec{S}$,$\vec{V}$,$\vec{O}$]  \\
  {Concat Addition} & {\newcite{Edelstein2016}} & [$\vec{S}$,$\vec{V}$] + [$\vec{V}$,$\vec{O}$] \\
  {Network} & {Ours} &  $\vec{S}\vec{V}^T$+$\vec{V}\vec{O}^T$+$\vec{S}\vec{O}^T$ \\ 
  \bottomrule
\end{tabularx}
  \caption{Composition functions used to obtain event vectors from function-specific vector spaces. $+$: addition, $\odot$: element-wise multiplication, $ \times $: dot product. [$\cdot$, $\cdot$]: concatenation.}
  \label{tab:compose}
\end{table}

\begin{table}[t!]
\def\arraystretch{1.0}
\small
\setlength\tabcolsep{3.5pt}
\begin{tabularx}{\linewidth}{llcc}
   \multicolumn{2}{c}{} & \multicolumn{2}{c}{\bf Spearman's $\rho$} \\
   \textbf{Model} & \textbf{Reference} & \textbf{GS199} & \textbf{KS108} \\
   \midrule
   {Copy Object W2V} & {\newcite{Milajevs2014}} &  \underline{0.46}& {0.66}  \\
   {Addition KS14} & {\newcite{Milajevs2014}} &  {0.28} & \underline{0.73} \\
   {} & {\newcite{Tilk2016}} & {0.34} & {-} \\
   {} & {\newcite{Weber2018}} & {-} & {0.71} \\
   \midrule
   \textbf{Ours: SVO d100} & {} & {} & {}\\
   {Verb only} & {Ours} &  {0.34} & {0.63} \\
   {Addition} & {Ours} &  {0.27} & \textbf{0.76} \\
   {Concat} & {Ours} &  {0.26} & \textbf{0.75} \\
   {Concat Addition} & {Ours} & {0.32} & \textbf{0.77} \\
   {Copy Object} & {Ours} &  {0.40} & {0.52} \\
   {Network} & {Ours} &  \textbf{0.53} & {-}  \\
\bottomrule

\end{tabularx}
\vspace{-1.5mm}
\caption{Results on the event similarity task. Best baseline score is \underline{underlined}, and the best overall result is provided in \textbf{bold}.}
\label{tab:event_similarity}
\end{table}

\bigskip
\noindent \textbf{Event Similarity.}
\label{subsec:eventsim}
We now test correlations of SVO-based event representations composed from a function-specific vector space (see Table~\ref{tab:compose}) to human scores in the event similarity task. A summary of the main results is provided in Table~\ref{tab:event_similarity}. We also report best baseline scores from prior work. The main finding is that our model based on function-specific word vectors outperforms previous state-of-the-art scores on both datasets. It is crucial to note that different modeling approaches and configurations from prior work held previous peak scores on the two evaluation sets.\footnote{Note the two tasks are inherently different. KS108 requires similarity between plausible triplets. Using the network score directly (which is a scalar, see Table~\ref{tab:compose}) is not suitable for KS108 as all KS108 triplets are plausible and scored highly. This is reflected in the results in Table~\ref{tab:event_similarity}.} Interestingly, by relying only on the representations from the $V$ subspace (i.e., by completely discarding the knowledge stored in $S$ and $O$ vectors), we can already obtain reasonable correlation scores. This is an indicator that the verb vectors indeed stores some selectional preference information as designed, i.e., the information is successfully encoded into the verb vectors themselves.


\bigskip
\noindent \textbf{Thematic-Fit Evaluation.}
\label{subsec:selpref}
Correlation scores on two thematic-fit evaluation data sets are summarised in Table~\ref{tab:thematic_fit}. We also report results with representative baseline models for the task: 1) a TypeDM-based model \citep{Baroni:2010}, further improved by \citet{Greenberg2015a,Greenberg2015b} (\textbf{G15}), and 2) current state-of-the-art tensor-based neural model by \citet{Tilk2016} (\textbf{TK16}). We find that vectors taken from the model trained in the joint 3-group SVO setup perform on a par with state-of-the-art models also in the 2-group evaluation on SV and VO subsets. Vectors trained explicitly in the 2-group setup using three times more data lead to substantial improvements on PADO414. As a general finding, our function-specific approach leads to peak performance on both data sets. The results are similar with 25-dim SVO vectors. 

Our model is also more light-weight than the baselines: we do not require a full (tensor-based) neural model, but simply function-specific word vectors to reason over thematic fit. To further verify the importance of joint multidirectional training, we have also compared our function-specific vectors against standard single-space word vectors \citep{Mikolov:2013nips}. The results indicate the superiority of function-specific spaces: respective correlation scores on MST1444 and PADO414 are 0.28 and 0.41 (vs 0.34 and 0.58 with our model). It is interesting to note that we obtain state-of-the-art scores calculating cosine similarity of vectors taken from \textit{two groups} found in the \textit{joint space}. This finding verifies that the model does indeed learn a joint space where co-occurring words from different groups lie close to each other. 



\bigskip
\noindent \textbf{Qualitative Analysis.}
We retrieve nearest neighbours from the function-specific ($S$, $V$, $O$) space, shown in Figure~\ref{fig:fs:motivation_nns}. We find that the nearest neighbours indeed reflect the relations required to model the SVO structure. For instance, the closest subjects/agents to the verb \textit{eat} are \textit{cat} and \textit{dog}. The closest objects to \textit{need} are three plausible nouns: \textit{help}, \textit{support}, and \textit{assistance}. As the model has information about group membership, we can also filter and compare nearest neighbours in single-group subspaces. For example, we find subjects similar to the subject \textit{memory} are \textit{dream} and \textit{feeling}, and objects similar to \textit{beer} are \textit{ale} and \textit{pint}.






\begin{table}[t]
\begin{minipage}[b]{1.0\linewidth}
\def\arraystretch{1.0}
\footnotesize
\begin{tabularx}{1.0\linewidth}{ll cc cc}
   \multicolumn{2}{c}{\textbf{Setup}} & \multicolumn{2}{c}{\bf Baselines} & \multicolumn{2}{c}{\bf Ours} \\
    \cmidrule(lr){1-2} \cmidrule(lr){3-4} \cmidrule(lr){5-6} 
    & &  &  & & \\
    & &  &  & {SVO} & {SV-VO}\\
   {Dataset} & {Eval} & {G15} & {TK16} & {(d=100)} & {(d=25)}\\
   \midrule
   {} & SV & {0.36} & {-}  & \textbf{0.37} & {0.31} \\
   {MST1444} & VO & {0.34} & {-}  &  \textbf{0.35} &  {0.35}  \\
   {} & full  & {0.33} & \textbf{\underline{0.38}} & {0.36} & {0.34} \\
   \midrule 
   {} &  SV  & \underline{0.54} & {-} &  {0.38} & \textbf{0.55} \\
   {PADO414} & VO & {0.53} & {-}  &  {0.54} & \textbf{0.61} \\
   {} & full  & \underline{0.53} & {0.52}  & {0.45} & \textbf{0.58}  \\
\bottomrule
\end{tabularx}
\caption{Results on the 2-variable thematic-fit evaluation. Spearman's $\rho$ correlation.}
\label{tab:thematic_fit}
\end{minipage}

\vspace{2mm}

\begin{minipage}[b]{1.0\linewidth}
\def\arraystretch{1.0}
\footnotesize
\begin{tabularx}{1.0\linewidth}{l XX XX}
   {} & \multicolumn{2}{c}{\bf async}  & \multicolumn{2}{c}{\bf sync }\\
   \cmidrule(lr){2-3} \cmidrule(lr){4-5}
   {} & {sep} & {shared} & {sep} & {shared} \\
   \cmidrule(lr){1-5} 
   \textbf{3 Variables} \\
   {KS108 Verb only} & {0.56} & {0.48} & {0.58} &  \textbf{0.60} \\ 
   {KS108 Addition} & {0.51} & {0.66} & {0.73} & \textbf{0.78} \\ 
   {GS199 Verb only} & {0.24} & {0.26} & {0.26} & \textbf{0.34} \\ 
   {GS199 Network} & {0.10} & {0.40} & {0.28} &  \textbf{0.52} \\ 
   \cmidrule(lr){1-5} 
   \textbf{2 Variables} \\
   {MST1444} & {0.17} & {0.10} & {0.30} & \textbf{0.39} \\ 
   {PADO414} & {0.41} & {0.21} &  \textbf{0.44} &  \textbf{0.44} \\ 
\bottomrule
\end{tabularx}
\caption{Evaluation of different model variants, by training regime and parameter sharing.}
\label{tab:async_vs_sync}
\end{minipage}
\vspace{-3mm}
\end{table}

\bigskip
\noindent \textbf{Model Variants.}
\label{sec:dis:sync}
We also conduct an ablation study that compares different model variants. The variants are constructed by varying 1) the training regime: asynchronous (\textit{async}) vs synchronous (\textit{sync}), and 2) the type of parameter sharing: training on separate parameters for each sub-network (\textit{sep})\footnote{With separate parameters we merge vectors from ``duplicate'' vector spaces by non-weighted averaging.} or training on shared variables (\textit{shared}). In the asynchronous setup we update the shared parameters per sub-network directly based on their own loss, instead of relying on the joint synchronous loss as in Section \ref{sec:model}. 

Table~\ref{tab:async_vs_sync} shows the results with the model variants, demonstrating that both aspects (i.e., shared parameters and synchronous training) are important to reach improved overall performance. We reach the peak scores on all evaluation sets using the \textit{sync}+\textit{shared} variant. We suspect that asynchronous training deteriorates performance because each sub-network overwrites the updates of other sub-networks as their training is not tied through a joint loss function. On the other hand, the synchronous training regime guides the model towards making updates that can benefit all sub-networks.

\section{Conclusion and Future Work}

We presented a novel multidirectional neural framework for learning function-specific word representations, which can be easily composed into multi-word representations to reason over event similarity and thematic fit. We induced a joint vector space in which several groups of words (e.g., S, V, and O words forming the SVO structures) are represented while taking into account the mutual associations between the groups. We found that resulting function-specific vectors yield state-of-the-art results on established benchmarks for the tasks of estimating event similarity and evaluating thematic fit, previously held by task-specific methods.

In future work we will investigate more sophisticated neural (sub-)networks within the proposed framework. We will also apply the idea of function-specific training to other interrelated linguistic phenomena and other languages, probe the usefulness of function-specific vectors in other language tasks, and explore how to integrate the methodology with sequential models. The pre-trained word vectors used in this work are available online at: \\
\url{https://github.com/cambridgeltl/fs-wrep}.

\section*{Acknowledgments}
This work is supported by the ERC Consolidator Grant LEXICAL: Lexical Acquisition Across Languages (no 648909) awarded to Anna Korhonen.

\bibliography{Biblio}
\bibliographystyle{acl_natbib}

\end{document}